\documentclass[letterpaper, 10pt, conference]{ieeeconf}      

\IEEEoverridecommandlockouts                              

\usepackage{graphics} 
\usepackage{epsfig} 
\usepackage{mathptmx} 
\usepackage{times} 
\usepackage{amsmath} 
\usepackage{amssymb}  
\usepackage{textcomp}
\usepackage{cite}
\usepackage{threeparttable}
\usepackage{booktabs}
\usepackage{multirow}
\usepackage{adjustbox}
\usepackage[bottom=57pt,top=54pt, left=54pt, right=54pt]{geometry}   
\usepackage{subcaption} 
\usepackage{xspace}

\newcommand{\etal}{\textit{~et~al.}\xspace}
\newcommand{\eg}{\textit{e.g.,}\xspace}
\newcommand{\ie}{\textit{i.e.,}\xspace}
\newcommand{\boldparagraph}[1]{\textbf{#1.}}

\usepackage[pagebackref,breaklinks,colorlinks]{hyperref}
\usepackage{relsize}

\title{\LARGE \bf
Reading the Room: Scale-ambiguous Semantic Registration of Sparse VSLAM Maps to Floor Plans in Partially Built Environments}

\author{Bjorn Stragier$^{1}$, Emilia Szyma\'nska$^{2}$,  J. Micah Prendergast$^{1}$ and Lukas Rosenberger Schmid$^{3}$%
\thanks{$^{1}$Cognitive Robotics Department, TU Delft, Delft, The Netherlands. {\tt\smaller bstragier, j.m.prendergast@tudelft.nl}}
\thanks{$^{2}$Hilti AG, Schaan, Liechtenstein. {\tt\smaller emilia.szymanska@hilti.com}}%
\thanks{$^{3}$ Machine Intelligence Group, University of Technology Nuremberg UTN,  Nuremberg, Germany. {\tt\smaller lukas.schmid@utn.de}
}%
}

\begin{document}

\maketitle
\thispagestyle{empty}
\pagestyle{empty}

\begin{abstract}
Many robotic construction applications, such as progress tracking, require registering a sensor trajectory to a prior 2D floor plan.
Visual capture is attractive for its low cost and availability, but aligning a visual simultaneous localization and mapping (VSLAM) map to a floor plan is difficult; VSLAM recovers geometry only up to an unknown similarity transform, the sparse point cloud is cross-modal to the symbolic floor plan, sites are only partially built during capture, and construction sites are visually repetitive, degrading map quality while removing distinctive features. 
Existing methods address at most two of these challenges.
We thus introduce \emph{Reading the Room} (RR), a global-local approach to register scale-ambiguous, sparse, semantic VSLAM maps to partially built floor plans.
Our key idea is to define a single robust, differentiable, class-balanced semantic-geometric energy function that governs all stages of the pipeline.
We develop two global initialization schemes, leveraging a single inaccurate user-provided correspondence if provided, or estimating the initialization globally using a cross-modal matching network.
Both schemes are designed to find initial candidates with respect to the energy function, which is then directly optimized for local refinement.
We evaluate RR on 33 sequences across three active construction sites against classical and learning-based baselines, achieving a 4.5× improvement in mean absolute trajectory error over the strongest baseline in the general unconstrained setting, which increases to 7.6× when an anchor is provided.
\end{abstract}


\section{INTRODUCTION}

Autonomous robots and perception technologies have the potential to transform modern construction. 
Tracking construction progress against design intent is one of the most direct levers to reduce project delays and cost overruns~\cite{Durdyev2020}. 
However, doing so requires registering a sensor trajectory captured by a robot or site operator to a prior 2D floor plan. 
To this end, camera-based capture is an especially attractive sensing solution given its vast availability, low cost, weight, and energy consumption, as well as providing visual fidelity and human interpretability.
While several works investigate registering dense LiDAR-based reconstruction to floor plans~\cite{Shahbandi2019, qiao2025lidar2bim, feng2023floorplannet}, the visual counterpart remains less explored.

\begin{figure}[t]
    \centering
    \includegraphics[width=0.48\textwidth]{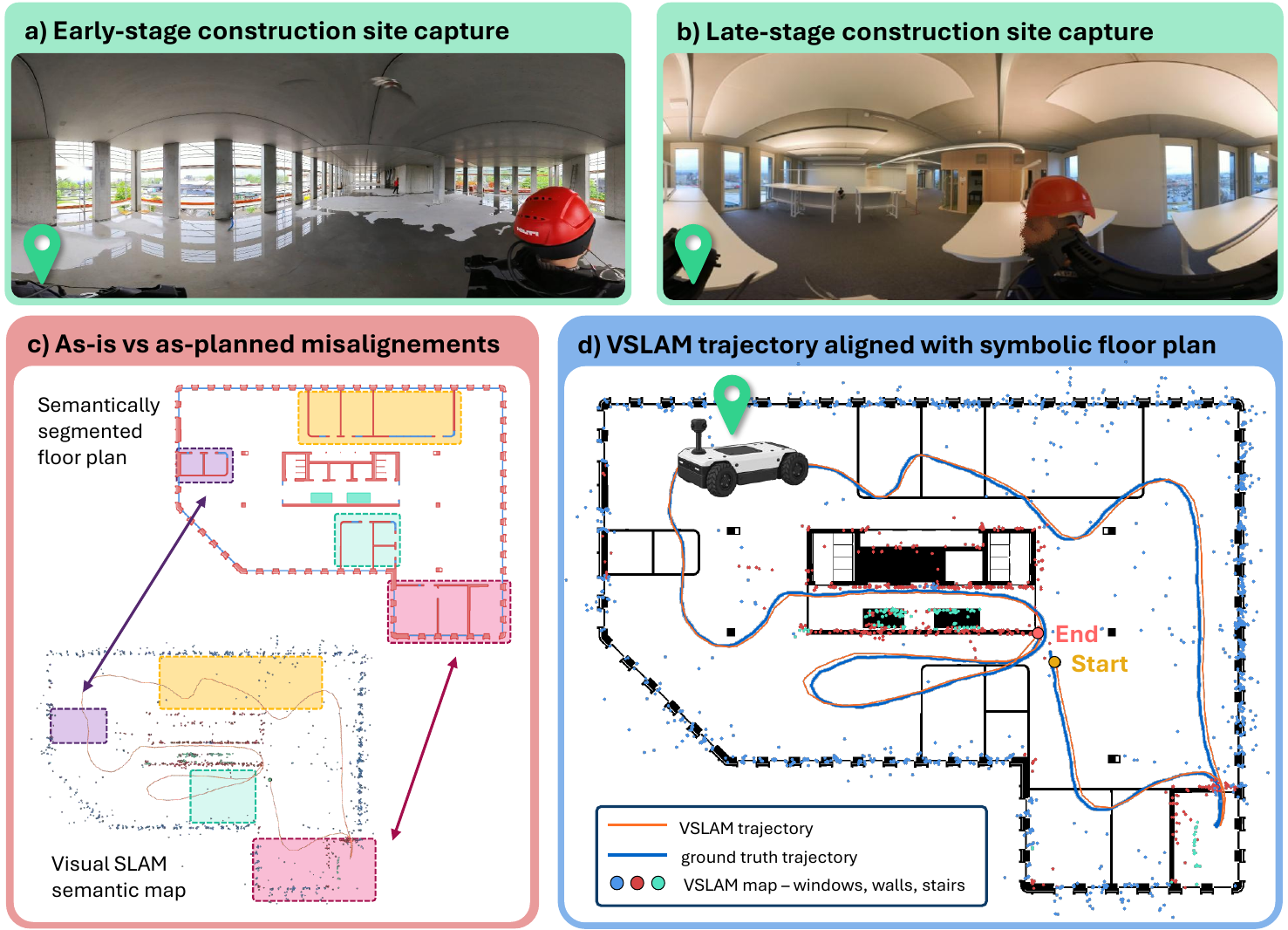}
    \caption{This work presents a method to align a noisy scale-ambiguous VSLAM trajectory to a floor plan.
    Our solution is able to operate in early construction phases (\textbf{a}) where notable parts of the floor plan may not yet have been built (\textbf{b}). This leads to missing structures in the sparse VSLAM map in addition to the cross-modality gap to the symbolic floor plan (\textbf{c}). Our method overcomes these challenges, achieving good trajectory alignment (\textbf{d}).}
    \label{fig:teaser}
\end{figure}

In this work, we study the task of aligning a sparse, scale-ambiguous 3D map, as recovered by a monocular VSLAM system~\cite{openvslam2019,ORBSLAM3_TRO}, to the floor plan of a site under active construction. 
Multiple properties compound to make this setting unusually difficult, as shown in Fig. \ref{fig:teaser}. 
First, monocular VSLAM recovers geometry only up to an unknown similarity transform, since scale is not observable without auxiliary sensing.
Second, the two representations are highly cross-modal, as VSLAM outputs a sparse, noisy 3D point cloud while the floor plan is a symbolic drawing of design intent. 
Third, construction sites are only partially built during capture, so large parts of the plan may not yet physically exist, making it an unreliable prior. 
Lastly, such sites are often visually repetitive, consisting mostly of walls, columns, and voids, which both degrades VSLAM quality and removes the distinctive features registration typically relies on. 

Existing approaches address at most one or two of these challenges. 
Cross-view indoor localization methods~\cite{boniardi19iros, howardjenkins2021lalaloc, howardjenkins2022lalaloc++, min2022laser, grader2025semrayloc, chen2024f3loc, meng2026discofloc, wueest2026unloc} generally assume known metric scale and a fully built environment. 
Methods embedding a floor plan directly into the SLAM pipeline~\cite{Gao_2022, zimmerman2023ral} similarly assume a known, already-built environment and a metric-range sensor such as LiDAR.
Map-alignment methods such as region decomposition~\cite{Shahbandi2019} and graph matching~\cite{feng2023floorplannet} require dense metric reconstructions a sparse VSLAM map cannot easily supply.
Recent sparse VSLAM methods instead by-pass dense reconstruction by detecting manually placed visual markers with positions marked a-priori in the floor plan~\cite{Bagh2026BIM360}. 
Learning-based cross-modal matchers~\cite{sun2021loftr,edstedt2024roma} can generalize across visually dissimilar modalities~\cite{he2025matchanything, ren2025minima}, but have never been evaluated on this domain. 
Thus, no existing method jointly handles unknown scale, sparse representations, the cross-modality gap, and partial construction.

To close this gap, we introduce \emph{Reading the Room} (RR), a global-local registration approach built around a single energy function. 
We propose a robust, differentiable, class-balanced semantic-geometric objective for cross-modal registration in partially built construction sites.
We then develop a two stage approach to globally initialize and locally refine this (or any other) energy function in feature-degraded and structurally ambiguous environments.
We thoroughly evaluate RR across 3 datasets of active construction sites spanning 33 total sequences. 
For practical relevance, we consider two cases, where an initial guess by a site operator is provided as well as the unconstrained global regsitration task, both of which are naturally handled by RR.
We compare RR against several classical and learning-based methods and achieve state-of-the-art performance in both tasks.
In summary, our main contributions are:

\begin{itemize}
    \item We introduce \textit{Reading the Room} (RR), a novel global-local approach to register scale-ambiguous sparse VSLAM maps against partially built symbolic floor plans.
    \item We propose a novel class-balanced semantic-geometric energy function and develop a two stage pipeline combining correspondence-free cross-modal dense feature matching or anchored initialization with local refinement to optimize this energy objective.
    \item We thoroughly evaluate RR across 33 sequences spanning three active construction sites, showing a $\geq4.5\times$ improvement in median ATE over the most competitive baselines, providing a first effective solution to this challenging problem.
\end{itemize}

\section{RELATED WORK}
\label{sec:related_work}

\subsection{Cross-view Localization}

Cross-view localization aims to localize a query image directly against a floor plan, without a SLAM algorithm or multi-view constraints in the loop.
Early work scored predicted room-layout edges against projected floor-plan edges~\cite{boniardi19iros}.
Embedding-based methods such as LaLaLoc~\cite{howardjenkins2021lalaloc} and its successors~\cite{howardjenkins2022lalaloc++, min2022laser} instead learn a shared panorama/floor-plan latent space.
More recently, methods render or predict depth and sometimes semantic rays~\cite{grader2025semrayloc} from candidate poses and compare them against observed depth~\cite{chen2024f3loc, meng2026discofloc, wueest2026unloc}.
These methods rest on several assumptions we cannot make.
First, depth-ray methods require a known pixel-to-metre ratio, whereas our floor plans carry no reliable scale. 
Second, we instead register a sparse map, which is easier to store, transmit, and anonymize than full single views.
Finally, all of these methods assume the built environment matches the plan closely, which may fail when large parts of a site are not yet built. 

\subsection{SLAM with Floor Plan Priors}

A separate line of work integrates a floor plan directly as a prior into a tracking or SLAM pipeline, rather than registering a map to the plan later.
FP-Loc~\cite{Gao_2022} tracks a 3D LiDAR from a known start pose against a floor plan by detecting ceiling/ground planes and matching wall features to floor-plan elements via an approximate nearest-neighbour field. 
Zimmerman\etal~\cite{zimmerman2023ral} extend this to global localization with a semantic map whose object-level cues resolve the structural symmetry via Monte Carlo localization. 
However, these methods assume a known, already-built environment and a metric-range sensor (LiDAR) that directly observes scale, neither of which holds in our setting.

\subsection{Map Registration}

Complementarily, map registration aims to estimate a rigid or similarity transform aligning a SLAM map to a floor plan.
Most prominently, point cloud methods such as ICP \cite{besl1992method} align dense LiDAR scans to reference geometry.
Recent works couple this with as-planned BIM/CAD models; diS-Graphs~\cite{shaheer2025tightly} jointly optimizes pose and a non-rigid plan deformation, while Lidar2BIM~\cite{qiao2025lidar2bim} registers a LiDAR map to a BIM model via a Hough-transform.
However, these methods assume a metric range sensor, whereas in our case scale is an unknown the algorithm must recover.
Alternatively, if scale is unknown but the modalities match, registration can be posed as a 2D image-alignment problem. Manual-correspondence methods sidestep the transform search entirely; Lu\etal~\cite{LU2021103723} use a 360-degree camera and VSLAM in construction but require several site-manager-selected tie points, while Bagh\etal~\cite{Bagh2026BIM360} require physical fiducial markers that are also manually annotated on the plan. 
The closest unconstrained approach is region decomposition~\cite{Shahbandi2019, Shahbandi2018Nonlinear, carpin2008fast}, which registers a dense occupancy grid map to a floor plan by scoring candidate region overlap.
A related graph-based work~\cite{feng2023floorplannet} matches regions with a learned graph network instead. 
All of these methods require a dense or volumetric reconstruction, which in a sparse, monocular VSLAM setting cannot reliably be produced.

\begin{figure*}
    \centering
    \includegraphics[width=\linewidth]{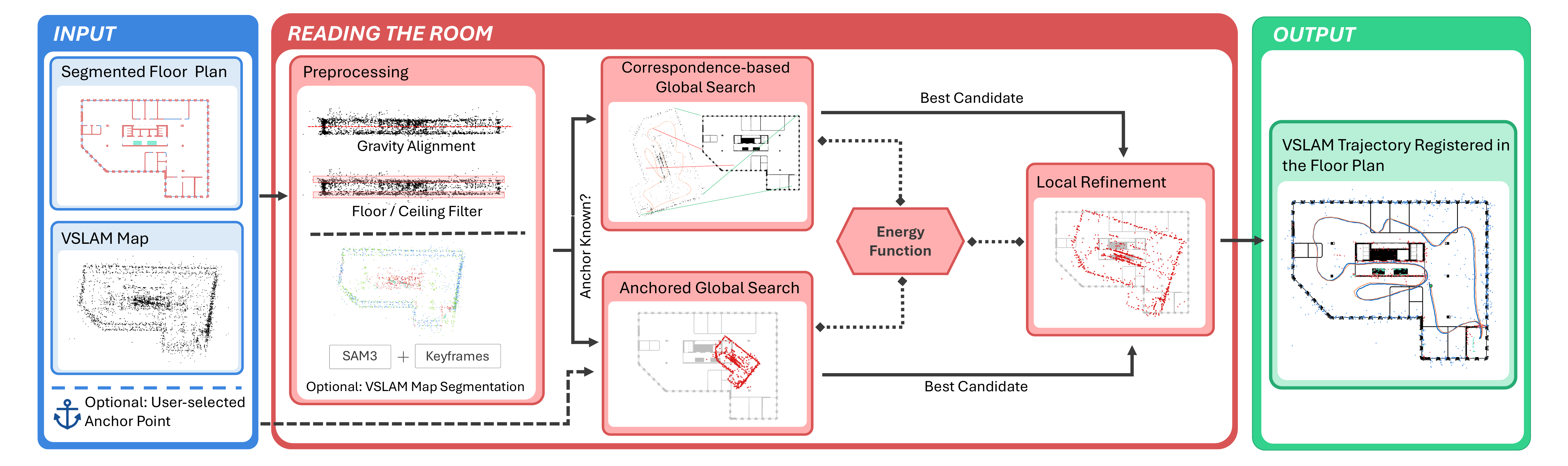} 
    \caption{Overview of \emph{Reading the Room}. To align a sparse VSLAM map to a symbolic floor plan, preprocessing first reduce the modality-gap by processing the 3D map into a BEV map (Sec.~\ref{sec:preprocessing}). A single energy function (Sec.~\ref{sec:cost}) guides the remaining steps. If a known correspondence is provided it is leveraged to constrain global initialization, otherwise global initialization is performed using dense cross-modal feature matching (Sec.~\ref{sec:global}). Finally, the estimate is refined (Sec.~\ref{sec:local}) to minimize the energy function locally.}
    \label{fig:methodology}
\end{figure*}  

\subsection{Cross-modal Feature Matching}

An alternative to registering full maps is to match local features between two images directly, without reducing either to an occupancy grid.
Classical descriptors such as SIFT~\cite{lowe2004distinctive} remain a common choice to match keypoints via handcrafted, scale- and illumination-invariant descriptors.
Learned sparse matchers instead match detected keypoints (typically from SuperPoint~\cite{detone2018superpoint}) via attention, as in SuperGlue~\cite{sarlin2020superglue} and LightGlue~\cite{lindenberger2023lightglue}. 
Detector-free dense matchers such as LoFTR~\cite{sun2021loftr}, its efficient variant~\cite{wang2024efficientloftr}, and RoMa~\cite{edstedt2024roma} regress correspondences over a dense grid, which tend to outperform sparse matchers precisely in the low-texture, repetitive conditions that also characterize construction sites.
All of these matchers are trained for same-modality correspondence and are not, by default, cross-modal. MatchAnything~\cite{he2025matchanything} and MINIMA~\cite{ren2025minima} address this gap without changing the underlying architecture by fine-tuning existing matchers on large synthetic cross-modal datasets (such as optical-to-infrared or optical-to-map imagery).
Both report that dense matchers retain their advantage over sparse and semi-dense ones even after cross-modal adaptation, suggesting cross-modal generalization is largely a property of training data rather than architecture. 
However, neither framework has yet considered the domain gap of sparse, unscaled VSLAM maps against a symbolic floor plan.

\section{PROBLEM STATEMENT}

We consider the problem of aligning a sparse 3D map $M$, as produced by a VSLAM system, to a symbolic floor plan $F$. 
The floor plan $F$ is given as a 2D image with semantic masks over structural semantic classes $\mathcal{C}$.
The map is given as $M=(P,T,C)$, where $P=\{p_i\}_{i=1}^{N_P}\in\mathbb{R}^3$ is a landmark point cloud, $T=\{T_j\}_{j=1}^{N_T}\in SE(3)$ is a sensor trajectory, and $C=\{c_i\}_{i=1}^{N_P}\subseteq\mathcal{C}\cup\{\varnothing\}$ assigns each landmark a semantic class $\mathcal{C}$, with $\varnothing$ denoting unlabelled landmarks. 
$M$ is expressed in an arbitrary frame of unknown scale, orientation, and translation relative to $F$. 
The goal is to recover the similarity transform $S^*\in Sim(3)$ mapping $M$ into the floor plan frame:
\begin{equation}
    S^{*}=\arg\min_{S\in Sim(3)}\lVert \pi(S\,T)-T_F^{GT}\rVert_2,
    \label{eq:problem_statement}
\end{equation}
where $\pi(\cdot)$ projects a transformed pose trajectory to its 2D position sequence in $F$ and $T_F^{GT}$ denotes the true 2D trajectory in plan frame. 

We consider two cases of this problem.
Optionally, a single known but inaccurate correspondence $(a,a_F)$ between one trajectory position $a\in\mathbb{R}^3$ and its user-marked pixel location $a_F\in\mathbb{R}^2$ on $F$ may be provided, \eg tapped once by a site manager at the start of capture. 
While no known correspondence is more general, such information is commonly available and practically relevant in construction settings.


\section{APPROACH}

An overview of our method, \textit{Reading the Room} (RR), is shown in Fig.~\ref{fig:methodology}. 
First, preprocessing (Sec.~\ref{sec:preprocessing}) transforms the provided map $M$ into a modality more similar to the floor plan $F$.
The central idea of our approach is to define a single energy function (Sec.~\ref{sec:cost}) that governs the coarse and fine registration steps.
To this end, a global initialization is estimated (Sec.~\ref{sec:global}) considering the initial correspondence if provided, or estimating it globally using cross-modal dense correspondence matching.
Finally, a local step (Sec.~\ref{sec:local}) refines the alignment of the map to the floor plan.


\subsection{Preprocessing}
\label{sec:preprocessing}

Since the map $M$ lives in a higher-dimensional space and represents a different modality than the floor plan $F$, we first aim to reduce this representation gap.
Since floor plans do not capture the vertical dimension, we remove this extra degree of freedom by estimating the direction of gravity using Principal Component Analysis (PCA) on the mean-centred landmark cloud.
Exploiting that most landmarks in built environments lie on the floor, ceiling, or wall surfaces, we find the eigenvector of least variance to be a robust proxy for verticality.
Similarly, while floor plans contain only structural elements such as walls, pillars, or stairs, the majority of map points typically lie on the floor and ceiling, which are not captured in $F$. 
To reduce this representation mismatch, floor and ceiling points are filtered out by retaining only the 15$^{\text{th}}$ to 85$^{\text{th}}$ percentiles of the height histogram. 
The remaining points are orthographically projected onto the horizontal plane, resulting in a shared bird's-eye-view (BEV) representation for both the map and plan. 
This reduces the unknowns from a full $Sim(3)$ estimation problem to $Sim(2)$, \ie finding $(s,\theta,t_x,t_y)\in Sim(2)$, a rotation about the vertical axis, an isotropic scale, and a 2D translation.
Finally, if no semantics are provided, a semantic class $c_i\in\mathcal{C}\cup\{\varnothing\}$ is estimated for each landmark by projecting it into its respective keyframes and querying a pre-trained vision-language segmentation model. We use SAM3~\cite{carion2025sam3segmentconcepts} with the class names in $\mathcal{C}$ as text prompts.
A landmark takes the class and confidence $w_i$ of the mask it projects into, and is left unlabelled $\varnothing$, contributing only to the class-agnostic term of Sec.~\ref{sec:cost} with confidence $w_i\geq0.5$, if it never projects into a mask.

\subsection{Energy Function}
\label{sec:cost}

The central idea of our approach is to define a single energy function which governs all other components of \textit{Reading the Room}.
This makes our architecture very flexible to adapt to other settings by defining other energy functions.
While many functions are admissible, we define the following semantic registration cost for our problem.
Since no explicit correspondences link the sparse landmark cloud to the floor plan, we score a candidate transform by how close each landmark lands to the nearest floor-plan structure of its own class.
For each class $c\in\mathcal{C}$, this proximity is the Euclidean distance to the nearest floor-plan pixel of class $c$.
For efficiency we pre-compute this once per floor plan as a distance transform $\mathrm{DT}_c$ over $F$'s class mask, reducing each candidate's score to a constant-time lookup. 
A candidate transform maps landmark $p_i$ to floor-plan pixel coordinates $q_i$, scored via bilinear interpolation of $d_i=\mathrm{DT}_{c(i)}(q_i)$ at every labelled landmark. 
To avoid not-yet-built structures influencing other landmarks, we only score the candidate transform with respect to the plan and not vice-versa.
Since landmarks are sparse and noisy, and construction sites often contain clutter, such as equipment or materials with no floor-plan counterpart, we truncate $d_i$ at $\tau$ such that outlier landmarks cannot dominate the cost, then aggregate the remainder with an additional Huber penalty $\rho_\delta$.
This cost is quadratic for informative gradients near good alignments but only linear for the larger residuals that are still common, while zeroing far-out estimates, thus keeping the objective well-behaved under gradient-based refinement (Sec.~\ref{sec:local}). 
Finally, because construction sites are often self-similar and most classes other than wall are rare, we average the confidence-weighted residual within each class independently and only then average across classes, giving a sparse but distinctive class (such as stairs) equal weight to more common ones.
The resulting energy function $\mathcal{L}_{\text{sem}}$ is thus of the form:

\begin{equation}
    \mathcal{L}_{\text{sem}}=\frac{1}{|\mathcal{C}|}\sum_{c\in\mathcal{C}}\left(\frac{\sum_{i:c(i)=c}w_i\,\rho_\delta(\min(d_i,\tau))}{\sum_{i:c(i)=c}w_i}\right).
    \label{eq:loss}
\end{equation}
Without notable parameter tuning, we choose $\tau=25\,$px and Huber threshold $\delta=10\,$px across all experiments. 


\subsection{Global Initialization}
\label{sec:global}

\boldparagraph{Anchored global search} 
When the anchor correspondence $(a,a_F)$ is available, it can be used to reduce the search space over the remaining transformation parameters. 
We parameterize the new points $q_i$ around the single known correspondence,
\begin{equation}
q_i=a_F+[t_x,t_y]+s\,R(\theta)(p_i-a),
\end{equation}
so that initializing $(t_x,t_y)$ at zero leaves only scale and rotation to be resolved.
Note that since our cost \eqref{eq:loss} is robust to local deviations, errors in the exact anchor position can well be absorbed. 
Furthermore, the full transform is later refined (Sec.~\ref{sec:local}), recovering small orientation and scale errors resulting from fixing the translation.

In general, registration costs, including \eqref{eq:loss}, are highly non-convex in $(s,\theta)$, since many rotations and scales bring some subset of landmarks into alignment with an incorrect match on the floor plan.
This challenge is exacerbated in partially built construction settings, as the number of distinct landmarks and classes is low, scenes are highly self-similar and thus susceptible to perceptual aliasing, room shapes are often similar and symmetric, and large parts of the plan may not yet have been built.
To address this, we perform a global search over $N=20{,}000$ random candidate scales and rotations, $\theta\sim\mathcal{U}[-180^\circ,180^\circ]$ and $s\sim\log\mathcal{U}[s_{\min},s_{\max}]$ with $s_{\min}=0.1$ and an adaptive upper bound
\begin{equation}
    s_{\max}=\eta\cdot\frac{d_F}{d_{\text{BEV}}},
    \label{eq:scale_max}
\end{equation}
where $d_F$, $d_{\text{BEV}}$ are the diagonals of the annotated floor-plan region and the BEV map, respectively, and $\eta=1.5$ is a safety factor to compensate for outlier measurements.
Each candidate is scored using our energy function $\mathcal{L}_{\text{sem}}$, and the lowest scoring candidate is selected for refinement.

\boldparagraph{Correspondence-based global search} 
When no anchor is available, an unconstrained global initial transform needs to be estimated.
We approach this through cross-modal dense feature matching. 
Our architecture is based on RoMa~\cite{edstedt2024roma}, however, we find that this matching network does not generalize well to our setting, even with large pre-trained cross-modal weights from MatchAnything~\cite{he2025matchanything}.
We instead propose a synthetic dataset to fine-tune RoMa~\cite{edstedt2024roma} initialized with MatchAnything~\cite{he2025matchanything} weights.

We generate synthetic data via floor plan augmentation. 
From only 100 labelled drawings we produce 2000 map--floor plan pairs. Each floor plan is cropped to a random per-axis fraction in $[0.6,1]$ to model partial trajectory coverage, then sparsified by retaining $[3\%,10\%]$ of foreground pixels to approximate landmark density. To simulate occluded, undetected, or not-yet-built structure, 1--10 rectangular patches ($5\%$--$25\%$ of the image each) are blacked out with probability 0.9. 
We model noise by adding spurious clutter points and adding Gaussian positional jitter to all points.
We apply random rotation $\theta \sim \mathcal{U}[-180^\circ,180^\circ]$, log-uniform scale in $[0.8,1.25]$, and translation up to $10\%$ of image size as the ground truth transform to recover. The clean floor plan mask always serves as the reference view during training.


At inference, since construction settings are highly self-similar, correspondences can be noisy and unreliable.
Thus, rather than relying on the accuracy of the network's identified correspondences, as standard RANSAC~\cite{fischler1981random} would, we instead generate several candidate transforms.
We retain the top $1{,}000$ correspondences by network confidence and fit a closed-form transform using Umeyama/orthogonal Procrustes~\cite{umeyama1991least} to $200$ two-point samples drawn randomly. 
To improve robustness, we also considers the correspondence-free parts of the floor plan by scoring every candidate with our energy function \eqref{eq:loss} applied to the full landmark point cloud, and retain the lowest-scoring candidate. 


\subsection{Local Refinement}
\label{sec:local}

Finally, starting from the initial guess obtained from Sec.~\ref{sec:global}, we jointly refine all four parameters $(s,\theta,t_x,t_y)$.
We directly optimize for the energy function \eqref{eq:loss} via bounded-memory quasi-Newton optimization~\cite{byrd1995limited}.
For efficient optimization, we use the closed-form gradient of $\mathcal{L}_{\text{sem}}$ and run for up to $5{,}000$ iterations. Parameter bounds are left unconstrained.

\section{Experiments}

\begin{table}[t]
\caption{Dataset Overview.}
\vspace{-5pt}
\centering
\setlength{\tabcolsep}{3pt}
\begin{adjustbox}{max width=\columnwidth}
\begin{tabular}{lcccccc}
\toprule
Site & Seq. & Floors & Loc. & Traj. [m] & Area [m$^2$] & Cov. [\%] \\
\midrule
Site A (Medium) & 19 & 9 & CH  & 134.8 & 1073.5 & 64.2 \\
Site B (Advanced) & 7  & 2 & USA & 564.5 & 2750.1 & 88.8 \\
Site C (Hard) & 7  & 3 & USA & 219.5 & 3974.7 & 28.7 \\
\bottomrule
\end{tabular}
\end{adjustbox}
\vspace{-5pt}
\begin{center}
\footnotesize
CH: Switzerland, USA: United States. Seq: number of sequences. Traj: average trajectory length. Area: average area spanned by the trajectory; Cov: average ratio of the floor plan area covered by the trajectory.
\end{center}
\label{tab:site_overview}
\end{table}

\begin{table*}[h!t]
\centering
\caption{Mean/median ATE [m] and recall at 1/3/5\,m ATE across all methods and datasets.}
\vspace{-5pt}
\label{tab:main_results}
\setlength{\tabcolsep}{2pt}
\begin{adjustbox}{max width=\textwidth}
\begin{tabular}{l cc cc cc cc}
\toprule
& \multicolumn{2}{c}{\textbf{Site A} (Medium)} & \multicolumn{2}{c}{\textbf{Site B} (Advanced)} & \multicolumn{2}{c}{\textbf{Site C} (Hard)} & \multicolumn{2}{c}{\textbf{Overall}} \\
\cmidrule(lr){2-3} \cmidrule(lr){4-5} \cmidrule(lr){6-7} \cmidrule(lr){8-9}
\textbf{Method} & ATE [m] $\downarrow$ & Recall [\%] $\uparrow$ & ATE [m] $\downarrow$ & Recall [\%] $\uparrow$ & ATE [m] $\downarrow$ & Recall [\%] $\uparrow$ & ATE [m] $\downarrow$ & Recall [\%] $\uparrow$ \\
\midrule
Oracle (performance lower bound) & 1.46/1.01 & 42.1/89.5/100.0 & 0.71/0.85 & 85.7/100.0/100.0 & 0.59/0.26 & 85.7/100.0/100.0 & 1.12/0.85 & 60.6/93.9/100.0 \\
\midrule
SIFT~\cite{lowe2004distinctive}+RANSAC~\cite{fischler1981random}$^*$ & 0.0/0.0 & 0.0/0.0/0.0 & 53.80/53.80 & 0.0/0.0/0.0 & 0.0/0.0 & 0.0/0.0/0.0 & 53.80/53.80 & 0.0/0.0/0.0 \\
RoMa~\cite{edstedt2024roma}+MatchAnything~\cite{he2025matchanything} & 13.94/16.12 & 10.5/26.3/31.6 & 40.63/42.41 & 0.0/0.0/0.0 & \textbf{42.15}/42.44 & 0.0/0.0/0.0 & 25.58/19.94 & 6.1/15.2/18.2 \\
Shahbandi and Magnusson~\cite{Shahbandi2019}$^\dagger$ & 18.93/18.18 & 0.0/0.0/0.0 & 45.53/44.99 & 0.0/0.0/0.0 & 0.0/0.0 & 0.0/0.0/0.0 & 29.57/24.94 & 0.0/0.0/0.0 \\
Reading the Room (ours) & \textbf{7.18}/\textbf{2.67} & \textbf{21.1}/\textbf{52.6}/\textbf{68.4} & \textbf{7.40}/\textbf{1.92} & \textbf{42.9}/\textbf{71.4}/\textbf{71.4} & 42.98/\textbf{35.04} & 0.0/0.0/0.0 & \textbf{14.82}/\textbf{4.34} & \textbf{21.2}/\textbf{45.5}/\textbf{54.5} \\
\midrule
ICP~\cite{besl1992method}+Anchor & 16.48/15.72 & 0.0/5.3/10.5 & 31.39/26.31 & 0.0/0.0/0.0 & 38.27/38.77 & 0.0/0.0/0.0 & 24.26/21.59 & 0.0/3.0/6.1 \\
Semantic ICP~\cite{besl1992method}+Anchor & 13.50/13.40 & 0.0/10.5/26.3 & 21.61/11.23 & 0.0/0.0/14.3 & 37.10/37.60 & 0.0/0.0/0.0 & 20.23/14.43 & 0.0/6.1/18.2 \\
Reading the Room (ours)+Anchor & \textbf{2.68}/\textbf{1.41} & \textbf{10.5}/\textbf{73.7}/\textbf{89.5} & \textbf{8.03}/\textbf{0.90} & \textbf{57.1}/\textbf{85.7}/\textbf{85.7} & \textbf{5.73}/\textbf{4.37} & 0.0/\textbf{14.3}/\textbf{57.1} & \textbf{4.46}/\textbf{1.89} & \textbf{18.2}/\textbf{63.6}/\textbf{81.8} \\
\bottomrule
\end{tabular}%
\end{adjustbox}
\vspace{-5pt}
\begin{center}
\footnotesize
Lower ATE and higher recall are better. Best result indicated in bold.  
$^*$SIFT~\cite{lowe2004distinctive} failed on 32/33 runs; the single successful run (Site B) is shown. 
$^\dagger$Shahbandi and Magnusson~\cite{Shahbandi2019} produced a valid transform on 10/33 runs (6 Site A, 4 Site B, 0 Site C), metrics are computed only over those runs. 
\end{center}
\end{table*}


\subsection{Experimental Setup}

\boldparagraph{Datasets}
We evaluate our method on 3 datasets, spanning 33 trajectories across 3 sites, 14 floors, and 2 countries. 
Sequences are captured on a custom device with synchronized LiDAR and 360\textdegree{} imagery.
The LiDAR points are used only to estimate the ground truth trajectory, which is manually aligned to the floor plan by an expert.
As a VSLAM solution, maps obtained from the equirectangular variant of OpenVSLAM~\cite{openvslam2019} are provided. 
The floor plans use the semantic class set $\mathcal{C} = \{wall,\, window,\, stairs\}$, where the corresponding map labels are obtained following Sec.~\ref{sec:preprocessing}.

Tab.~\ref{tab:site_overview} summarizes the three datasets.
While all datasets are challenging, we order the three sites in increasing difficulty to better assess method performance and limitations. 
Site A (Medium) has the the smallest floors (1073.5\,m$^2$ on average) and contains comparatively few distinct rooms per plan. 
Site B (Advanced) has larger floors (2750.1\,m$^2$) but also a relatively high coverage (88.8\%).
Site C (Hard) has the largest floor area (3974.7\,m$^2$) and the lowest coverage (28.7\%).
In addition, the covered area exhibits structural symmetry, such that multiple candidate alignments can appear plausible to a geometry- or structure-based approach. 

\boldparagraph{Metrics} 
Following \eqref{eq:problem_statement}, we evaluate trajectory localization using the absolute trajectory error (ATE):
\begin{equation}
    \mathrm{ATE}=\sqrt{\frac{1}{N}\sum_{j=1}^{N}\lVert \hat t_j - t_j^{GT}\rVert_2^2},
    \label{eq:ate}
\end{equation}
for a sequence with $N$ keyframes, where $\hat t_j = \pi(S^{*}T_j)\in\mathbb{R}^2$ denotes the estimated position at keyframe $j$ after applying the recovered transform $S^{*}$ and $t_j^{GT}$ denotes the corresponding ground-truth position.
Since failed alignment can produce outlier errors, we report the mean and median ATE across all sequences.
In addition, following indoor localization tasks~\cite{min2022laser, wueest2026unloc}, we report recall as the success rate where ATE falls below a given threshold $\tau \in \{1\text{m}, 3\text{m}, 5\text{m}\}$.

\boldparagraph{Baselines}
As outlined in Sec.~\ref{sec:related_work}, we are not aware of any methods directly addressing the problem at hand.
We therefore compare against several of the most closely related baselines. 
For the correspondence-based global registration, we compare against SIFT~\cite{lowe2004distinctive}+RANSAC~\cite{fischler1981random} as a classical feature-based method, and  RoMa~\cite{edstedt2024roma} with pretrained weights from MatchAnything~\cite{he2025matchanything} representing the state-of-the-art in learnt cross-modal feature matching.
Most closely related to ours, we further compare against the region decomposition approach of Shahbandi and Magnusson~\cite{Shahbandi2019}, which was specifically designed for map-to-floor-plan alignment, albeit for denser maps than are typically obtainable by VSLAM.
When an initial anchor is provided, we compare against Iterative Closest Point (ICP)~\cite{besl1992method} and its semantic variant (Semantic ICP). For a fair comparison, we augment both variants to initialize with the same anchor correspondence as our method and iteratively refit scale, rotation, and translation via closed-form Umeyama/orthogonal Procrustes~\cite{umeyama1991least} on nearest-neighbor correspondences.



\subsection{Trajectory Localization Performance}

\begin{figure*}[t]
    \centering
    \includegraphics[width=\textwidth]{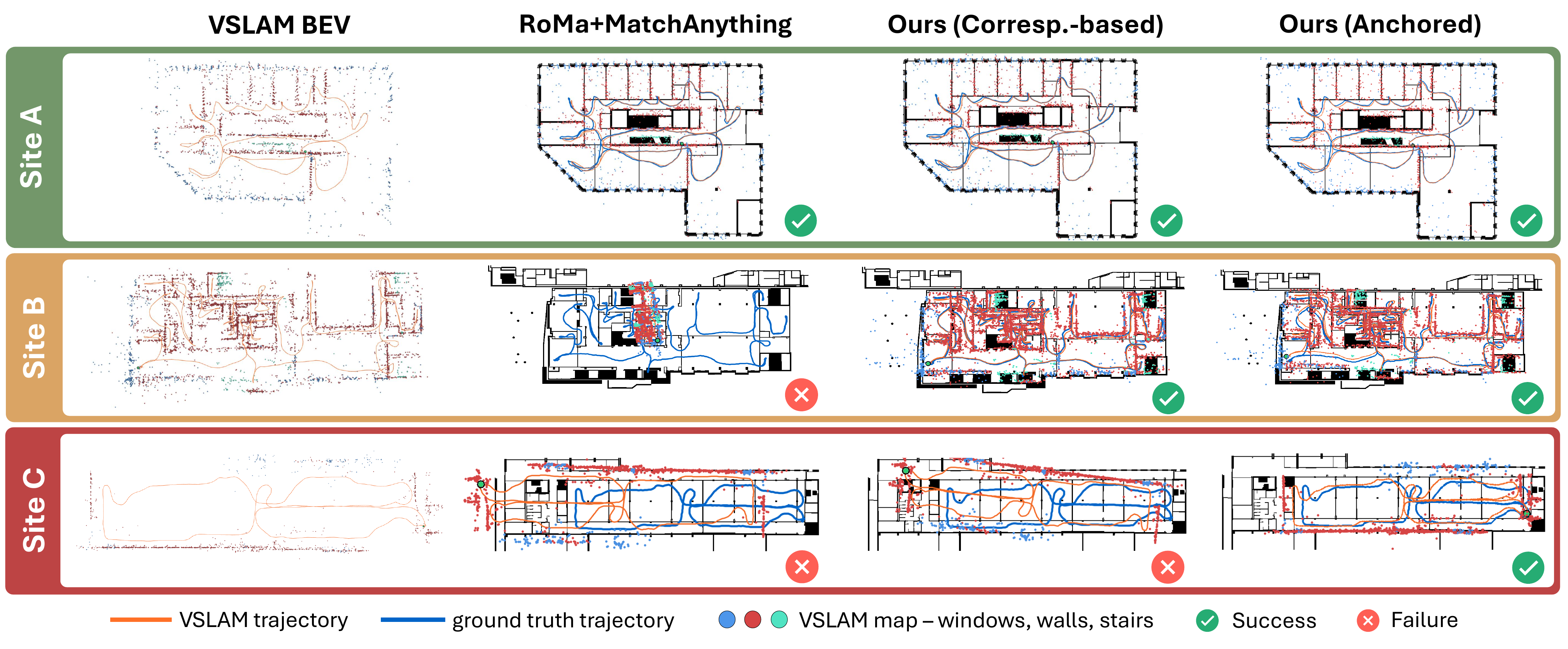}
    \caption{Qualitative registration results for one representative sequence per site, increasing in difficulty from top to bottom.
    VSLAM BEV maps are rotated and scaled for visualization; this does not reflect their initial condition for registration.
    On Site A, all three methods recover visually similar trajectories, consistent with the small floor area, high coverage, and unique structural layout of the site.
    In Site B, the RoMa+MatchAnything collapses the entire map into several rooms. This yields a low cost as the outer walls align with the room walls, highlighting the challenge of scale-ambiguous registration.
    In Site C, both RoMa+MatchAnything and our approach fail due to the symmetry of the scene and align the sparse observations flipped. 
    If a single noisy user-provided anchor is available, our method can leverage this to recover the right symmetry and achieve a good solution in spite of orientation and scale ambiguity.}
    \label{fig:qualitative_results}
\end{figure*}

The performance of all methods across all datasets is shown in Tab.~\ref{tab:main_results}. 
First, we observe that the maps obtained from VSLAM, in spite of loop closure and global optimization, are not perfectly drift free, adding to the challenges of the registration task.
To quantify this, we report an oracle performance by directly fitting the trajectory to the ground truth path via Umeyama/orthogonal Procrustes~\cite{umeyama1991least}. 
This residual error of approximately 1m on average is thus a lower bound on the performance that can be achieved.
Given the difficulty of the task at hand, we leave non-rigid registration and floor-plan conditioned trajectory deformation, required to eliminate the final residual error, for future work.

\boldparagraph{Correspondence-based Global Methods}
We observe that SIFT~\cite{lowe2004distinctive}+RANSAC~\cite{fischler1981random} fails on 32 of 33 runs, as the modality gap between VSLAM map and floor plan is too large for classical feature matching to bridge.
RoMa~\cite{edstedt2024roma}+MatchAnything~\cite{he2025matchanything} is the strongest prior baseline and manages to find solutions for all cases. 
However, performance degrades sharply with building complexity. 
This is most evident in the recall dropping to 0 for Site B and Site C.
Qualitative results in Fig.~\ref{fig:qualitative_results} illustrate some of the challenges inherent in our problem setting. 
While the matching performance is high on Site A, challenges such as self-similarity of the rooms in Site B lead to local minima where the entire plan collapses into smaller rooms, or the symmetry in Site C leading to a result that is 180$^\circ$ flipped.
This suggests that some pretrained cross-modal correspondences can be found but only yield usable results when the site is simple and visually distinct enough to resolve scale and rotation ambiguities without further adaptation, which is rarely the case in construction settings.
Shahbandi and Magnusson~\cite{Shahbandi2019} has been designed for this case, however, their region matching approach requires a dense reconstruction for optimal performance.
In our setting, this produces a valid transform only on 10/33 runs, highlighting the challenge of having to operate on sparser landmarks. 
This is particularly pronounced in Site C, where the site coverage of only 28.7\% contains too little boundary structure for their assumptions to hold, leading to a 0\% success rate.
In contrast, our approach achieves significant improvements, reducing mean ATE by 72\% compared to the most competitive baseline.
Most notably, while outliers fail catastrophically and drive up mean ATE, our energy function for global matching and refinement yields many more close results, manifesting in a 359\% improvement in median ATE and significantly higher recall.
While our method achieves the best results for Site A and is the only method to succeed in Site B, we also fail to correctly register Site C, underscoring the difficulty of the task and need for further research.

\boldparagraph{Anchored Methods}
Tab.~\ref{tab:main_results} shows that providing a single anchoring correspondence guess can significantly improve registration performance. 
Especially in Site C, where all correspondence-based global methods failed, an initial guess is crucial in anchoring the partial observations, allowing RR to achieve a median ATE of only 4.37m and 57\% recall@5m.
However, it becomes evident that an initial guess alone is not sufficient to solve the problem given scale and rotation ambiguity, as the ATE for ICP remains high. 
Similarly, while semantics alone improve the ICP performance, there remains a notable performance gap. 
In contrast, our robust and class-balanced global energy function used for candidate scoring and refinement achieves practically relevant results across most sites. 
This is also reflected in the qualitative results in Fig.~\ref{fig:qualitative_results}, where the provided anchor is able to resolve the symmetry issue in Site C, although the scale is not quite perfect yet.

\subsection{Ablation Studies}

We ablate several of the contributions and design choices of our approach to isolate their effectiveness and demonstrate their importance. 
The results are aggregated in Tab.~\ref{tab:ablations}.

\begin{table}[t]
\centering
\caption{Mean/median ATE [m] $\downarrow$ for different ablations.}
\vspace{-5pt}
\label{tab:ablations}
\setlength{\tabcolsep}{3pt}
\begin{adjustbox}{max width=\columnwidth}
\begin{tabular}{l l cc cc cc cc}
\toprule
Study & Method & Site A  & Site B  & Site C & Overall\\

\midrule
\multirow{3}{*}{\rotatebox{90}{\parbox{0.95cm}{\centering \smaller Energy\\Function}}} &
Geometric & 17.77/19.63 & 47.24/57.04 & 55.61/59.22 & 32.05/21.93 \\
& Semantic & \textbf{7.14}/3.34 & 17.64/2.00 & \textbf{41.74}/38.28 & 16.71/5.13 \\
& Ours &  7.18/\textbf{2.67} & \textbf{7.40/1.92} & 42.98/\textbf{35.04} & \textbf{14.82/4.34} \\

\midrule
\multirow{2}{*}{\rotatebox{90}{\parbox{0.65cm}{\centering \smaller Global\\Init.}}} &
Uniform Grid   & 10.15/10.08 & 45.47/48.48 & 37.66/40.97 & 23.48/15.44  \\
& Ours & \textbf{2.67/1.65} & \textbf{7.94/0.90} & \textbf{6.22/4.69}  & \textbf{4.54/1.89} \\

\midrule
\multirow{4}{*}{\rotatebox[origin=c]{90}{\parbox{1.3cm}{\centering \smaller Local\\Refinement}}} &
Anchored Coarse & 3.02/2.24 & 8.22/1.37  & 6.56/5.16 & 4.87/2.32 \\
& + Refinement & \textbf{2.67/1.65}  & \textbf{7.94/0.90} & \textbf{6.22/4.69} & \textbf{4.54/1.89} \\
\cmidrule(lr){2-6}
& Global Coarse & 6.99/2.94 & 10.80/1.82 & 43.98/41.43  & 15.64/4.91\\
& + Refinement &  \textbf{6.68/2.73}  & \textbf{10.10/0.94}& \textbf{41.72/36.36}& \textbf{14.84/3.91}\\

\midrule
\multirow{2}{*}{\rotatebox{90}{\parbox{0.65cm}{\smaller Anchor}}} &
User-marked & 2.68/1.41 & 8.03/0.90 & 5.73/4.37 & 4.46/1.89\\
& Ground-truth  & 2.86/1.81 & 1.22/0.78 & 4.73/3.02& 2.91/1.81  \\

\midrule
\multirow{2}{*}{\rotatebox{90}{\parbox{0.65cm}{\smaller Match}}} &
RANSAC~\cite{fischler1981random} & 9.94/6.97 &  14.42/6.29 & 54.79/61.32 & 20.40/12.23 \\
& Ours & \textbf{7.18/2.67}  & \textbf{7.40/1.92} & \textbf{42.98/35.04} & \textbf{14.82/4.34} \\

\bottomrule
\end{tabular}%
\end{adjustbox}
\end{table}


\boldparagraph{Energy Function}
We isolate the contribution of our robust class-balanced semantic-geometric cost in the global registration setting by ablating each component of \eqref{eq:loss}.
We use only our robust geometric distance term (Geometric), adding the semantic weights (Semantic), as well as the additional class-imbalance aware aggregation (Ours). 
While the semantic information yields the largest improvement, each component further raises performance across all sites.
This demonstrates the importance of not only having robust costs for partially built scenes but also accounting for the self-similarity and scarcity of semantic landmarks in construction.


\boldparagraph{Importance of the Global Initialization Scoring}
To test whether refinement alone can recover a correct alignment, we anchor the map at the same user-selected position and utilize the same energy function, but replace our global initialization strategy (Sec.~\ref{sec:global}) with a deterministic grid search, where instead of sampling and pre-scoring candidates every grid point is refined and the best result is kept. 
We use the same distribution as for our search for the grid and select  32 orientations $\times$ 32 scales for 1024 total initializations. 
The results in Tab.~\ref{tab:ablations} demonstrate the highly complex energy landscape of this registration problem, where even for a dense uniform grid many solutions end up in suboptimal local minima. 
While our approach requires more samples, pre-scoring the candidates with our energy function \emph{before} refinement makes this an efficient approach.



\boldparagraph{Importance of Local Refinement}
While the previous experiments showed the importance of the global stage, we here analyse the importance of the local refinement.
To this end, we run both global stages, \ie with and without an anchor provided, both with and without the final refinement stage. 
While we observe that the global stage, \ie finding the right local minimum attraction basin, dominates the performance, our unified energy function proves to be effective also in driving the estimate toward the exact local minimum and consistently improves the median ATE by $\sim$24\% on average.

\boldparagraph{Robustness to Anchor Inaccuracy}
We compare our anchored pipeline under the noisy, user-marked anchor used throughout all experiments in this paper against providing the ground-truth anchor.
While the more accurate anchor tends to yield better performance, our approach overall drops only 4.4\% in median ATE, demonstrating the robustness of our method to inaccurate anchor placement.
This is achieved through our two stage global scoring + local refinement approach, where our robust energy function subsumes additional inaccuracies.
Robustness to anchor placement by real construction site operators further suggests the practical relevance of our approach.

\boldparagraph{Candidate Scoring vs. Correspondence Matching}
As a final study, we compare our approach to score a range of candidate transforms provided by the global matching network (Sec.~\ref{sec:global}) using our energy function against optimizing the number of matching correspondences using RANSAC~\cite{fischler1981random}. 
While RANSAC is effective at finding the largest inlier set, because the landmarks present in partial construction settings are so self-similar and the scenes are structurally similar or symmetric, the largest inlier set does not yield the optimal transform in this case. 
In comparison, our method evaluates a number of the most confident correspondences predicted by the network and scores them globally using our energy function, which yields a significantly improved estimate that can be passed on to local refinement.

\subsection{Runtime}

\begin{table}[t]
\centering
\caption{ Runtime [s] $\downarrow$ as mean $\pm$ std per pipeline stage.}
\vspace{-5pt}
\label{tab:runtime_by_site}
\setlength{\tabcolsep}{2pt}
\renewcommand{\arraystretch}{0.92}
\begin{adjustbox}{max width=\columnwidth}
\scriptsize
\begin{tabular}{l l cccc}
\toprule
 & & Site A & Site B & Site C & Overall \\
\midrule
\multicolumn{2}{l}{Preprocessing*} & 0.46 $\pm$ 0.07 & 0.99 $\pm$ 0.11 & 0.44 $\pm$ 0.10 & 0.57 $\pm$ 0.24 \\
\midrule
\multirow{2}{*}{\rotatebox{90}{\parbox{0.55cm}{\centering \smaller Global\\Init.}}} &
Anchored     & 6.95 $\pm$ 1.84 & 15.23 $\pm$ 1.99 & 5.07 $\pm$ 1.92 & 8.31 $\pm$ 4.15 \\
& Corr.-based$^\dagger$  & 0.81 $\pm$ 0.23 & 0.84 $\pm$ 0.03 & 0.74 $\pm$ 0.03 & 0.80 $\pm$ 0.18 \\
\midrule
\multirow{2}{*}{\rotatebox{90}{\parbox{0.55cm}{\centering \smaller Loc.\\Refine.}}} &
Anchored     & 0.08 $\pm$ 0.02 & 0.10 $\pm$ 0.03 & 0.06 $\pm$ 0.02 & 0.08 $\pm$ 0.03 \\
& Corr.-based & 0.11 $\pm$ 0.05 & 0.20 $\pm$ 0.05 & 0.08 $\pm$ 0.03 & 0.12 $\pm$ 0.06 \\
\midrule
\multirow{2}{*}{\rotatebox{90}{\parbox{0.55cm}{\centering \smaller Total}}} &
\textbf{Anchored}     & \textbf{7.49 $\pm$ 1.90} & \textbf{16.33 $\pm$ 2.08} & \textbf{5.57 $\pm$ 1.99} & \textbf{8.96 $\pm$ 4.39} \\
& \textbf{Corr.-based} & \textbf{1.38 $\pm$ 0.26} & \textbf{2.02 $\pm$ 0.14} & \textbf{1.26 $\pm$ 0.11} & \textbf{1.49 $\pm$ 0.36} \\
\bottomrule
\end{tabular}%
\end{adjustbox}
\vspace{-3pt}
\begin{center}
    \footnotesize *Preprocessing does not include SAM3 semantic segmentation. \\$^\dagger$Mean network inference time across all runs is $0.66 \pm 0.17$s.
\end{center}
\end{table}

The runtime of each stage of Reading the Room is shown in Tab.~\ref{tab:runtime_by_site}. 
All experiments are run on an Intel Core i9-10900K CPU, with the matching network running on an NVIDIA RTX 3090 GPU.
Overall, we note that our efficient implementation achieves registration results within only a few seconds.
We observe significantly faster runtimes for the correspondence-based global initialization. 
While the network inference accounts for the largest share of $\sim$75\%, far fewer candidates are scored compared to the anchored approach, making it $\sim6\times$ faster overall.
However, this gap is not uniform across sites, as the correspondence-based runtime stays comparatively stable regardless of scene content while the anchored search scales with the number of observed landmarks, growing on the larger more densely covered Site B.
Inversely, the local refinement is faster for the anchored approach, as fewer iterations are needed.
Nonetheless, the runtime cost is dominated by the global initialization stage.
This highlights a practical trade-off between the two strategies, where the anchored pipeline's accuracy (Tab.~\ref{tab:main_results}) comes at an increased and less predictable runtime cost.
In summary, we believe that with only few seconds of CPU runtime our approach is practical for robot localization or operation by a site manager.
\section{CONCLUSION}

We presented Reading the Room, a global-local method for registering scale-ambiguous, sparse VSLAM maps to partially built floor plans. At its core is a novel robust, class-balanced semantic-geometric energy function that scores candidate transforms without requiring correspondences, remains robust to unbuilt structure and outlier landmarks, and weights rare but distinctive classes equally with common ones. 
This same function drives global initialization, using either anchored search or cross-modal matching, as well as local refinement.
Across 33 sequences over three active construction sites, our solution outperforms classical and learning-based baselines, achieving a 4.5$\times$ or greater improvement in median ATE over the strongest prior baseline and a median ATE below 2m when an anchor is provided. Ablations confirm that each component contributes to this performance, and that RR is robust to anchor inaccuracy.
Nonetheless, registration remains unreliable on our hardest site, where low coverage and symmetry drive all methods into incorrect minima. 
As our oracle analysis reveals a residual error of approximately 1 m as a result of VSLAM drift, future work should explore non-rigid, floor-plan-conditioned trajectory deformation.
Similarly, further mechanisms to resolve symmetric ambiguities without manual anchoring and correspondence matching in self-similar environments are promising avenues for future research.

\bibliographystyle{IEEEtran}
\bibliography{references}

\end{document}